\documentclass[letterpaper, 10 pt, conference]{ieeeconf}  

\IEEEoverridecommandlockouts                              

\usepackage{graphics} 
\usepackage{amsmath} 
\usepackage{amssymb}  
\usepackage{xcolor}
\usepackage{cite}

\title{\LARGE \bf
SSL-Interactions: Pretext Tasks for Interactive Trajectory Prediction
}

\author{Prarthana Bhattacharyya$^{1}$, Chengjie Huang$^{2}$ and Krzysztof Czarnecki$^{3}$
\thanks{$^{1}$Prarthana Bhattacharyya, $^{2}$Chengjie Huang and $^{3}$Krzysztof Czarnecki are with the Faculty of Engineering, University of Waterloo, 200 University Avenue, Waterloo, ON,
Canada. Email: ({\tt\small p6bhatta@uwaterloo.ca}, {\tt\small c.huang@uwaterloo.ca}, {\tt\small k2czarne@uwaterloo.ca})
}%
}

\makeatletter
\let\NAT@parse\undefined
\makeatother
\usepackage[colorlinks=true, citecolor=blue]{hyperref}
\usepackage[capitalize]{cleveref}
\crefname{section}{Sec.}{Secs.}
\Crefname{section}{Section}{Sections}
\Crefname{table}{Table}{Tables}
\crefname{table}{Tab.}{Tabs.}

\DeclareUnicodeCharacter{2212}{-}
\DeclareUnicodeCharacter{2212}{+}
\usepackage{graphicx}
\usepackage{algorithm}
\usepackage{algpseudocode}

\usepackage{pifont}
\newcommand{\xmark}{\ding{55}}%
\newcommand{\cmark}{\ding{51}}%

\usepackage{tabularx}
\usepackage{booktabs}
\usepackage{bm}
\usepackage[free-standing-units=true]{siunitx}
\usepackage{multirow}
\DeclareUnicodeCharacter{2225}{-}
\pdfoutput=1

\begin{document}

\maketitle
\thispagestyle{empty}
\pagestyle{empty}

\begin{abstract}
This paper addresses motion forecasting in multi-agent environments, pivotal for ensuring safety of autonomous vehicles. Traditional as well as recent data-driven marginal trajectory prediction methods struggle to properly learn non-linear agent-to-agent interactions. We present SSL-Interactions that proposes pretext tasks to enhance interaction modeling for trajectory prediction. We introduce four interaction-aware pretext tasks to encapsulate various aspects of agent interactions: range gap prediction, closest distance prediction, direction of movement prediction, and type of interaction prediction. We further propose an approach to curate interaction-heavy scenarios from datasets. This curated data has two advantages: it provides a stronger learning signal to the interaction model, and facilitates generation of pseudo-labels for interaction-centric pretext tasks. We also propose three new metrics specifically designed to evaluate predictions in interactive scenes. Our empirical evaluations indicate SSL-Interactions outperforms state-of-the-art motion forecasting methods quantitatively with up to 8\% improvement, and qualitatively, for interaction-heavy scenarios.
\vspace{-0.2cm}
\end{abstract}

\section{Introduction}
\label{sec:intro}
Anticipating future behavior of surrounding agents is crucial for safety of autonomous vehicles. By accurately predicting the future trajectories of dynamic traffic agents, autonomous vehicles can strategically plan and execute maneuvers avoiding collisions. However, in a multi-agent environment, numerous entities simultaneously interact and influence each other's behaviors. The main challenge lies in developing models that accurately captures these complex interactions between agents. 

Early trajectory prediction methods utilized techniques such as the Kalman filter and model-based approaches \cite{ModelBased}. These methods focused on learning the dynamic actions of \textit{individual} agents. Additionally, models like Social Forces \cite{SocialForces} were used to capture human-like behaviors for goal navigation. However, these hand-crafted methods cannot handle complex interactions among multiple agents in real-world scenarios. Data-driven methods have significantly advanced the field of multi-agent prediction, leveraging large datasets to generate more accurate and contextually aware predictions \cite{Lanegcn}. However, when it comes to highly interactive scenarios, these marginal predictions can sometimes produce unrealistic results \cite{M2I}. This is because the interactions between agents, which can significantly influence their future trajectories, are not adequately considered in marginal prediction models. 
On the other hand, we can use a joint predictor \cite{M2I} to predict trajectories in a joint space over multiple agents to generate trajectories that are more compliant with the actual scene. A joint predictor captures the interactions between agents more accurately by jointly considering their states in the prediction process. However the prediction space increases exponentially with the number of agents, resulting in a high computational cost that can make this approach impractical for real-time applications. This underlines the need for a \textit{compromise between marginal predictions and joint predictions} - an approach that can capture the interactions between agents without incurring the high computational cost of joint prediction. 
\begin{figure*}
    \centering
    \includegraphics{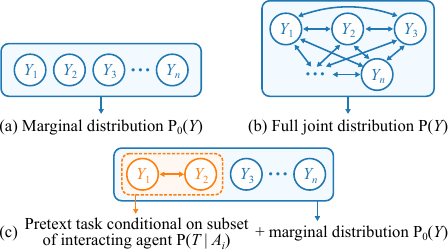}
    \caption{Comparison of approaches for training multi-agent forecasting systems. (a) \textit{Interactions between agents are not considered explicitly.}  Each node represents an agent's state, and arrows denote the aggregation of these marginal distributions $\text{P}_0(\boldsymbol{Y})$. (b) \textit{Interactions between all pairs of agents are considered.} The nodes represent the random variables, and the bidirectional edges between every pair of nodes denote the dependencies among all pairs of random variables. $P(\boldsymbol{Y})$ indicates the estimation of the full joint interaction distribution.  (c) \textit{SSL-Interactions considers interactions between pseudo-labeled pairs of agents.} The nodes represent the random variables, grouped by interaction-specific dependencies in subset $A_i$ enclosed by dashed-line rectangles. $P(\boldsymbol{T}|A_i)$ represents the conditional distributions for pretext task. This is used to train the interaction-module in a self-supervised setup. $P_0(\boldsymbol{Y})$ captures the marginal distributions of all the agents.}
   \label{intro_pic}
   \vspace{-0.4cm}
\end{figure*}

In this paper, we propose to utilize pretext tasks for interaction modeling in motion forecasting. This offers several advantages. First, pretext tasks offer a way to decompose the joint distribution into a conditional and a marginal, presenting a computationally feasible yet effective solution, as shown in \cref{intro_pic}. Second, pretext tasks enable the incorporation of interaction-specific domain knowledge during the learning process. By designing tasks that reflect meaningful aspects of agent interaction, we can guide the model to capture relevant patterns in the data, which acts as an effective regularization technique. Third, pretext tasks can be designed in a manner that allows pseudo-labels to be generated from the available data itself. This enables us to leverage large amounts of un-annotated data. We introduce four interaction-aware pretext tasks for motion forecasting: range gap prediction, closest distance prediction, direction of movement prediction, and type of interaction prediction. 

In addition, we propose an approach to curate interactive scenarios from a large dataset and explicitly label interaction pairs. This curation process not only provides stronger training signal for the interaction model, it also generates effective pseudo-labels that facilitates pretext task learning. 

To evaluate the effectiveness of our proposal, we introduce three metrics that provide a better assessment of predictions in interactive scenarios. These metrics include interactive min-FDE(i-min-FDE), non-interactive min-FDE(ni-min-FDE), and collision awareness metric(CAM). 
We also demonstrate the benefits of SSL-Interactions over the state-of-the-art, in terms of quantitative metrics and qualitative analysis. 

Our contributions are threefold: 
\begin{itemize}
    \item We propose a framework called SSL-Interactions, that leverages pretext tasks to improve interaction modeling for motion forecasting. Specifically, we develop four pretext tasks, designed to capture various aspects of interaction based on domain-specific knowledge.
    \item We propose a simple way to curate interaction-specific scenarios from datasets and to explicitly label pairs of interacting agents within a given scenario. This approach is crucial for generating pseudo-labels for our interaction-centric pretext tasks, and provide a stronger learning signal to the interaction component.
    \item Through empirical evaluation, we demonstrate that our proposed framework can surpass a state-of-the-art motion forecasting method both quantitatively, with up to an 8\% improvement, and qualitatively. Furthermore, we introduce three new metrics specifically designed to evaluate predictions within interactive scenes.
\end{itemize}
\section{Related Work}
\label{sec:relatedwork}
\subsection{Marginal prediction} In recent years, approaches driven by data have shown superior results as they learn interactions directly from input data, particularly in the context of real-world driving scenarios. To predict interactive pedestrian trajectories in crowded scenes, Social-LSTM\cite{SocialLSTM} and Social-GAN\cite{SocialGAN} leverage social pooling mechanisms that effectively capture social influences from neighboring agents. These mechanisms enable the model to incorporate the collective behavior of nearby individuals when making trajectory predictions. Graph neural networks (GNNs) possess robust relational inductive biases and have exhibited remarkable performance in tasks that involve relational reasoning, including visual question answering and complex physical systems \cite{RN}. 
GNNs have been extensively used in traffic scenarios to model interactions between agents \cite{STGCNN, SpaGNN, ILVM, Trajectron++}. A specialized form of attention mechanism, which is a special case of GNNs, that allow the model to weigh the importance of different entities in the scene relative to each other, is used in \cite{SocialBiGat, Scenetransformer, Lanegcn, Multiplefutures}. GNNs implicitly model agent-to-agent interactions by structuring the data as a graph, where each node represents an agent and edges signify the relationship or interaction between agents. Through propagation and update rules, GNNs can capture the influence of one agent on another during prediction, effectively encoding the interactive dynamics \textit{implicitly} within the traffic scenario. However, this prediction is marginal because it's made independently for each agent, considering only the current state of its neighbors in the graph. It does not account for the joint interaction effects that could arise from the simultaneous movement of multiple agents.

\subsection{Joint prediction} Joint prediction of trajectories in a traffic scene, while providing a comprehensive view of the interactions between multiple agents, comes with its own set of challenges. It requires modeling the dependencies between all agents, which can increase the computational complexity exponentially with the number of agents, making it impractical for real-time applications. \cite{MultiplexAttention} proposes utilizing \textit{explicit} latent interaction graphs via multiplex attention to infer high-level abstractions for improved multi-agent system forecasting. The M2I model \cite{M2I} leverages a marginal predictor to produce predictive samples for the 'influencer' agents, and employs a conditional predictor to project the future trajectories of the 'reactor' agents based on the influencer's anticipated path. This approach, however, requires the training of a relation predictor capable of discerning the influencer and reactor roles even during the inference stage. This may introduce noise and potentially struggle to scale effectively when dealing with multiple interactive agents.

\subsection{Prediction with Pretext Tasks} The use of pretext tasks has enhanced the modeling of interactions and relational reasoning across different contexts. \cite{TellMeWhy} demonstrates that predicting language descriptions and explanations as an auxiliary task can significantly enhance reinforcement-learning (RL) agents' abilities to infer abstract relational and causal structures in complex environments. \cite{LanguagePretext} introduces a trajectory prediction model that utilizes linguistic intermediate representations to enhance forecasting accuracy and model interpretability. \cite{DomainPseudoLabels} uses pseudo-labels in a multi-agent trajectory prediction setting to induce an informative, interactive latent space for a conditional variational auto-encoder (CVAE), thereby mitigating posterior collapse and improving the trajectory prediction accuracy. Social-SSL \cite{SocialSSL} is the closest to our approach, and utilizes self-supervised pre-training to improve data efficiency and generalizability of Transformer networks in trajectory prediction. However, the proposed pretext tasks do not necessarily have a direct impact on the downstream motion forecasting task. This is because success in these pretext tasks isn't tightly coupled with improved performance in motion forecasting. 

To the best of our belief, we are the first to introduce SSL-Interactions, a framework that factorizes and scales joint distribution prediction into marginal prediction and pretext task distribution prediction. The pretext tasks, informed by domain-specific knowledge, are closely linked to the performance of downstream motion forecasting.

\begin{algorithm}[t]
\caption{Label Interacting Trajectories} \label{chap-6-algorithm1}
\begin{algorithmic}[1]
\Procedure{LabelInteractingTrajectories}{$\mathcal{T}$}
    \State \textbf{Input:} Scene Trajectories $\mathcal{T} = \{\mathcal{Y}^{GT}_1, ... , \mathcal{Y}^{GT}_n\}$
    \State \textbf{Output:} Filtered interacting pairs $\mathcal{I}$
    \For{each pair of trajectories $(\mathcal{Y}^{GT}_i, \mathcal{Y}^{GT}_j)$}
        \State Calculate $d_{ij} = min(||p_i^{t_1} - p_j^{t_2}||)$ 
        \If{$d_{ij} < d_{th}$}
            \State Mark pair $(\mathcal{Y}^{GT}_i, \mathcal{Y}^{GT}_j)$ as interacting
        \EndIf
    \EndFor

    \For{each target agent $t$}
        \State Classify target agent's intent as $I_t$
    \EndFor

    \For{each interacting pair $(\mathcal{Y}^{GT}_i, \mathcal{Y}^{GT}_j)$}
        \If{$I_t \in \{\text{`Left-Turn'}, \text{`Left-Turn-Waiting'}\}$}
            \State Keep oncoming agents
        \Else
            \State Remove oncoming agents
        \EndIf
    \EndFor

    \State \textbf{return} Filtered interacting pairs $\mathcal{I}$
\EndProcedure
\end{algorithmic}
\end{algorithm}
\begin{figure}[t]
  \centering
   \includegraphics[height=6cm]{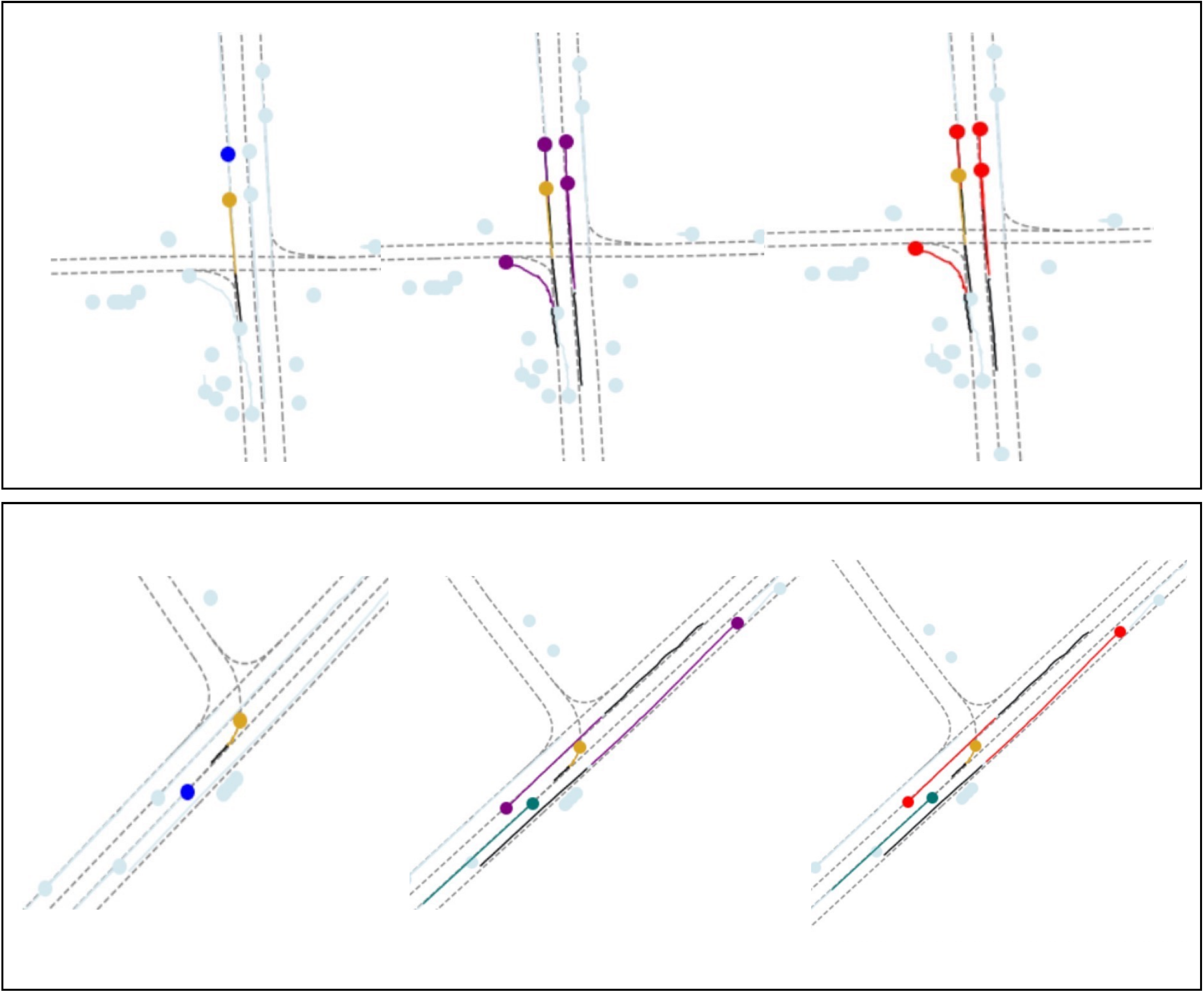}
 \caption[Illustration of the proposed data curation method for explicitly labeling pairwise interactions]{Illustration of the proposed data curation method for explicitly labeling pairwise interactions. The future trajectory of the target agent is denoted in yellow, while the past inputs are given in black. The first step involves identifying agents within a specified distance threshold, indicated by the violet color. Nonetheless, only distance thresholding is inadequate, as vehicles moving in opposite directions frequently do not interact. In the second step, oncoming agents are filtered out if the target agent intends to proceed straight, but are retained if the target's intended action is a left turn. The final interacting agents' future trajectories are given in red.}
   \label{chap-6-dataset_curation:left and straight}
\vspace{-0.4cm}
\end{figure}
\begin{figure*}[t]
  \centering
   \includegraphics[width=0.85\textwidth]{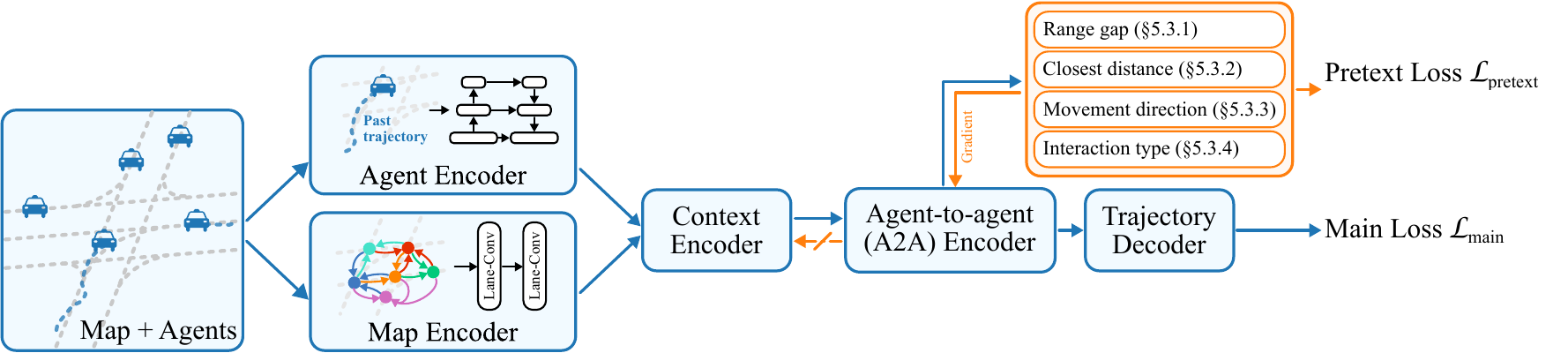}
    \caption{Schematic diagram of the proposed model incorporating a pretext task as an auxiliary loss. The agent encoder processes each agent's observed trajectory, while the map encoder handles high-definition (HD) map encoding. These representations are initially passed to a context encoder, generating map-conditioned agent features. These are subsequently processed by an agent-to-agent attention-based encoder, which encodes inter-agent dependencies. This comprehensive representation informs both the future trajectory decoder and the proposed pretext task component. The pretext task loss, benefiting from a stop gradient, exclusively trains the agent-to-agent encoder, ensuring only interaction-specific features are harnessed by the pretext tasks. Consequently, any improvements can be specifically attributed to enhanced interaction modeling within the agent-to-agent encoder. The pretext task loss, serving as an auxiliary task, is discarded during the inference phase.}
   \label{architecture}
\end{figure*}

\section{Problem Formulation}
\label{sec:chap-6-problem}
\label{sec:chap-6-notations}
We are given as input a dataset of past movements for $N$ actors. Each actor is represented by a set of $(x, y)$ coordinates indicating their center locations during the preceding $T_p-1$ time steps. To prepare the data for analysis, we preprocess each trajectory into a sequence of displacements denoted by $\mathcal{S}_i = \{\Delta \boldsymbol{s}_i^{−\{T_p-1\}+1}, . . . , \Delta \boldsymbol{s}_i^{−1}, \Delta \boldsymbol{s}_i^{0}\}$. Here, $\boldsymbol{s}_i^{t_p}$ represents the 2D displacement from the time-step $t_p - 1$ to $t_p$. This pre-processing ensures that the model concentrates on forecasting relative coordinates rather than absolute positions. The set of past motions of all N actors is given by $\mathcal{M} = \{\mathcal{S}_1, ..., \mathcal{S}_N\}$. Additionally, we have access to a high-definition (HD) map that includes lane and semantic attributes. We use a scene-centric embedding where we use an agent of interest’s position as the origin, and encode all roadgraph and agents with respect to it. The map elements are encoded into an embedding $\boldsymbol{h}_0$. The ground-truth future motion of each actor in the scene is also provided and denoted by $\mathcal{Y}^{GT}_i = \{(x^1_i , y^1_i ),..., (x^{T_c}_i , y^{T_c}_i )\}$  over a prediction horizon of $T_c$. We can also write this as $\mathcal{Y}^{GT}_i = \{p_i^1, p_i^2, ..., p_i^{T_c}\}$ and $\mathcal{Y}^{GT}_j = \{p_j^1, p_j^2, ..., p_j^{T_c}\}$. $||p_i^{t_1} - p_j^{t_2}||$ denotes the Euclidean distance between point $p_i^{t_1}$ from trajectory $\mathcal{Y}^{GT}_i$ and point $p_j^{t2}$ from trajectory $\mathcal{Y}^{GT}_j$. The set of ground-truth futures is given by $\mathcal{O}_{\text{GT}} = \{\mathcal{Y}^{GT}_i | i = 1, ..., N\}$. The future trajectory prediction for agent $i$ is represented by $\mathcal{Y}_i = \{(\hat{x}^1_i , \hat{y}^1_i ),..., (\hat{x}^{T_c}_i, \hat{y}^{T_c}_i)\}$ and $\mathcal{O}_{\text{pred}} = \{\mathcal{Y}_i | i = 1, ..., N\}$. But there could be multiple feasible future predictions for a given past input. Our goal is to predict $K$ possible future trajectories,  $\mathcal{P}_F$, where $\mathcal{P}_F = \{\mathcal{O}_{\text{pred, 1}}, \mathcal{O}_{\text{pred, 2}}, ... , \mathcal{O}_{\text{pred, K}}\}$. We assume that the agents interacting in a given scene are not provided with specific interaction labels. Thus we want a trajectory prediction model that learns to model the distribution $p({\boldsymbol{\mathcal{Y}}}|\mathcal{M}, \boldsymbol{h}_0)$.

\section{SSL-Interactions}
\label{sec:SSL-Interactions}
We introduce SSL-Interactions, a trajectory prediction framework that leverages self-supervised learning to capture social interaction from our designed pretext tasks. SSL-Interactions trains on both the main task of trajectory prediction and the self-supervised pretext tasks simultaneously. 
The overall architecture is illustrated in \cref{architecture}.
\par \textit{Advantages:} Unlike transformers, which require large amounts of data to learn effectively, SSL-Interactions can be trained on smaller datasets by leveraging the pretext tasks. By generating self-supervised labels related to the downstream task, we can use a larger amount of training data to train the data representation, rather than relying solely on the downstream task's labels. Additionally, this approach ensures that the model learns useful features relevant to the main task.
\subsection{Labeling Interactions}
\label{sec:chap-6-dataset curation}
In \cref{sec:chap-6-problem}, our assumption is that agents in a given scene do not have access to explicit interaction labels. To train a predictor that utilizes pretext tasks, we first produce a dataset that contains explicit interaction labels. 
\par To ensure effective interaction modeling, we visually query a random subset of the data and discover several limitations. In some cases, the target agent is the only vehicle present in a large area around it, limiting the potential for interaction with surrounding vehicles. In other cases, the target agent cannot be influenced by surrounding vehicles due to factors such as distance or the absence of spatio-temporal conflict.
It is clear that a model trained solely on this data would not contribute significantly to interaction modeling. Therefore, to improve our model's ability to capture interactions, it is necessary to curate a dataset that specifically addresses these limitations and includes instances where meaningful interactions occur. We propose a simple but effective approach to identify interacting pairs of trajectories based on the spatial distance between the agents and their intents. The overall algorithm is given in \cref{chap-6-algorithm1}. For every scene, we simplify our problem to filter only $(\mathcal{Y}^{GT}_{\text{target}}, \mathcal{Y}^{GT}_j)$. This is because datasets like Argoverse \cite{Argoverse} have already pre-selected the target agent of interest, and we can focus our attention solely on the interactions between the target agent and the other agents in the scene. Thus $A_i$ for agent $i$ is the output of \cref{chap-6-algorithm1}. 
A visual illustration of the data curation process is given by \cref{chap-6-dataset_curation:left and straight}. Further details about the algorithm are in \cref{appendix}.
\subsection{Proposed Pretext Tasks}
\label{sec:chap-6-proposed pretext task}
\begin{figure*}[ht]
  \centering
  \includegraphics[width=0.5\textwidth]{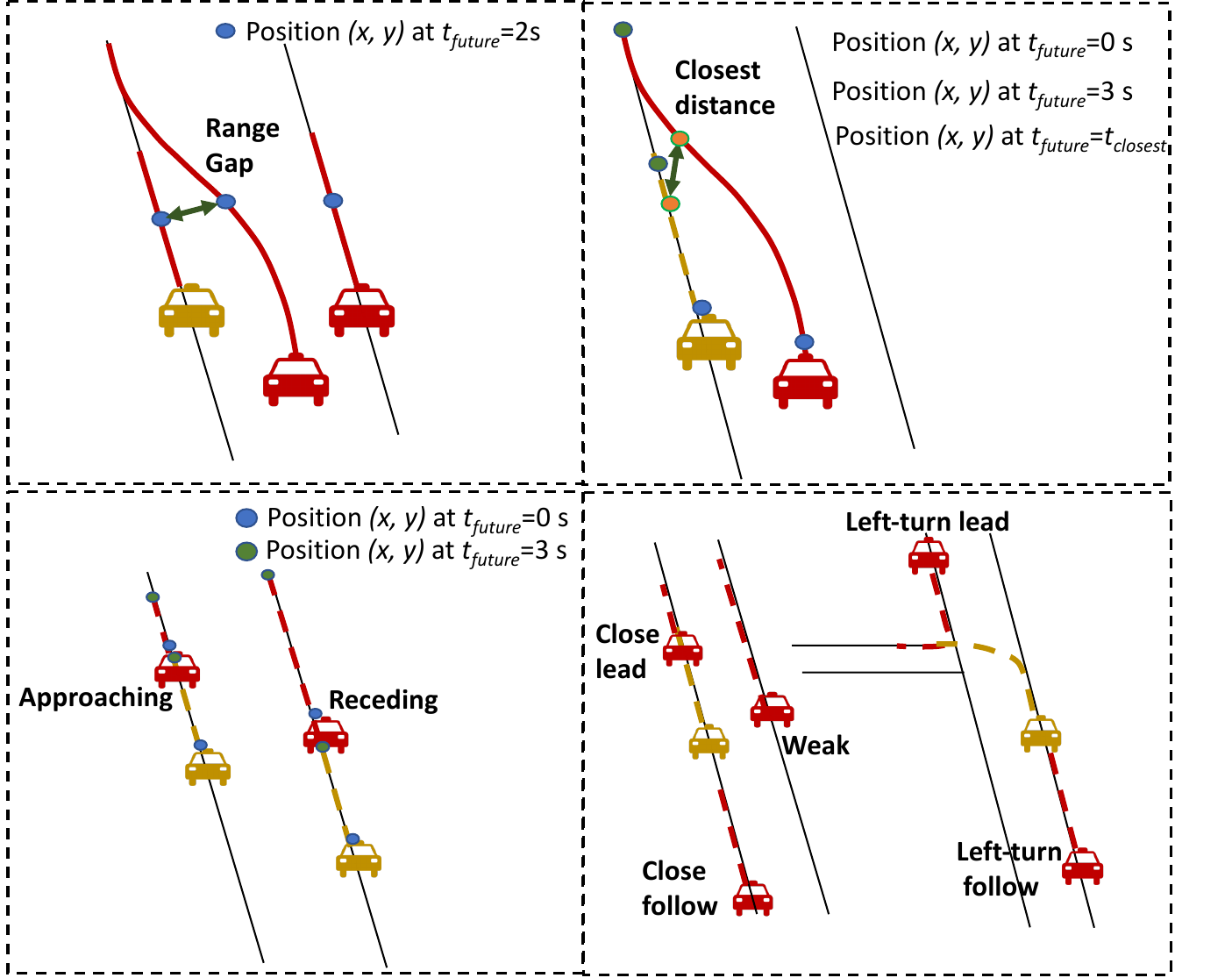}
 \caption[Proposed interaction-based pretext tasks]{Our proposed interaction-based pretext tasks, range-gap prediction(top-left), closest-distance prediction(top-right), direction of movement prediction(bottom-left) and type of interaction prediction(bottom-right), as described in \cref{sec:chap-6-proposed pretext task}.}
   \label{chap-6-pretext_tasks}
   \vspace{-0.4cm}
\end{figure*}

In this section, we present four pretext tasks for interaction and motion modeling:
1) range-gap prediction, 2) closest distance prediction, 3) direction of movement prediction and 4) type of interaction prediction. The overall setup of the pretext task is given by \cref{chap-6-pretext_tasks}.
\subsubsection{\textbf{Range-gap Prediction}}
In this pretext task, the main idea is to predict the range-gap between pairs of agents at a future time step $t=2s$, given their past trajectory information. The range-gap can be defined as the difference in the distance traveled by two agents during a specific time interval. This task focuses on learning the interaction patterns between agents by considering the spatial and temporal correlations in their trajectories.

Let the pretext task feature extractor for this task to be denoted by $f_{\text{RG}}$ with the parameters $\boldsymbol{\Theta}_{\text{RG}}$. We consider the ground-truth range gap between two agents $i$ which is the target agent and $j$ to be $||p_{\text{target}}^{t_1} - p_j^{t_2}||$ at time $t_1=2s$ and $t_2=2s$. The auxiliary range gap feature predictor takes the following inputs: the agent-to-agent interaction feature from the standard motion forecasting architecture as described in \cref{sec:chap-6-standard model} 
represented by $\tilde{\boldsymbol{v}}_{\text{target}}$ and $\tilde{\boldsymbol{v}}_{j}$, and the distance between them at $t_p=0$ given by $d^0_{\text{target}:j}$. The output is a predicted range gap distance between the target agent and $j$.
 \begin{equation}\label{chap-6-eq:7} 
\small
{d}_j^{\text{RG}} = f_{\text{RG}}(\tilde{\boldsymbol{v}}_{\text{target}} -  \tilde{\boldsymbol{v}}_j, d^0_{\text{target}:j})
\normalsize
\end{equation}
 The objective function $\boldsymbol{\mathcal{L}}_{\text{RG}}$ for guiding the range-gap prediction is formulated using Smooth-L1 loss as shown in \cref{chap-6-eq:8}. $k_\text{target}$ is the number of interacting agents with respect to the target agents as obtained from \cref{sec:chap-6-dataset curation}.
\begin{equation}
\label{chap-6-eq:8} 
\small
\boldsymbol{\mathcal{L}}_{\text{RG}}(\boldsymbol{\Theta}_{\text{RG}}) = \frac{1}{k_\text{target}} \sum_{j=1}^{k_\text{target}}{L_\text{reg}}({d}_j^{\text{RG}}, ||p_{\text{target}}^{t_1} - p_j^{t_2}||)
\normalsize
\end{equation}
 
\par \textit{Benefit to final forecasting performance:} By predicting range-gaps, the model learns to represent and capture the dependencies between the agents' movements more effectively. For example, if the range gap predicted between two agents is low, it could indicate that the two agents are getting closer to each other and may potentially collide. Conversely, if the range gap predicted is high, it could indicate that the two agents are moving away from each other and may not interact. This helps the model understand the underlying structure and patterns of agent interactions better, ultimately leading to more accurate motion forecasts.

\subsubsection{\textbf{Closest-distance Prediction}}
In this pretext task, the main idea is to predict the minimum distance between two agents in the future. In other words, we directly predict the closest distance that will occur between the agents at corresponding time steps in the $T_c$ time step future, without explicitly analyzing the distance at each future time step. This task enables the model to acquire meaningful representations of agent interactions in dynamic environments. 

Let the pretext task feature extractor for this task to be denoted by $f_{\text{CD}}$ with the parameters $\boldsymbol{\Theta}_{\text{CD}}$. We consider the ground-truth closest distance between two agents $i$, which is the target agent, and $j$ to be calculated as: $D_j^{\text{GT}} = \text{min}\{D_{\text{target}:j}(t) | t \in {1, ..., T_c}\}$, where, $D_j^{\text{target}:j}(t) = ||p_{\text{target}}^{t} - p_j^{t}||$. Additionally, we want to setup this task as a classification task instead of a regression task. Thus we choose four classes with labels $y_j^{\text{CD}}$: 
\begin{equation}\label{chap-6-eq:9} 
\small
y_j^{\text{CD}} = 
\begin{cases}
      0 & \text{if} \quad 0 < D_j^{\text{GT}} \leq 5 \\
      1 & \text{if} \quad 5 < D_j^{\text{GT}} \leq 10 \\
      2 & \text{if} \quad 10 < D_j^{\text{GT}} \leq 15 \\
      3 & \text{if} \quad D_j^{\text{GT}} > 15 \\
    \end{cases}     
\normalsize
\end{equation}

The auxiliary closest distance feature predictor takes the following inputs: the agent-to-agent interaction feature from the standard motion forecasting architecture \cite{Lanegcn} as described in \cref{sec:chap-6-standard model} 
represented by $\tilde{\boldsymbol{v}}_{\text{target}}$ and $\tilde{\boldsymbol{v}}_{j}$, and the distance between them at $t_p=0$ given by $d^0_{\text{target}:j}$. The output is a predicted closest distance between the target agent and $j$. 
 \begin{equation}\label{chap-6-eq:10} 
\small
{d}_j^{\text{CD}} = \text{Softmax} (f_{\text{CD}}(\tilde{\boldsymbol{v}}_{\text{target}} -  \tilde{\boldsymbol{v}}_j, d^0_{\text{target}:j}))
\normalsize
\end{equation}

The objective function $\boldsymbol{\mathcal{L}}_{\text{CD}}$ for guiding the closest distance prediction is formulated using cross-entropy loss as shown in \cref{chap-6-eq:11}. $k_\text{target}$ is the number of interacting agents with respect to the target agents as obtained from \cref{sec:chap-6-dataset curation}.
\begin{equation}\label{chap-6-eq:11} 
\small
\boldsymbol{\mathcal{L}}_{\text{CD}}(\boldsymbol{\Theta}_{\text{CD}}) = \frac{1}{k_\text{target}} \sum_{j=1}^{k_\text{target}}{L_\text{cross-entropy}}({d}_j^{\text{CD}}, y_j^{\text{CD}})
\normalsize
\end{equation}

\textit{Benefit to final forecasting performance:} Estimating the closest distance between agents enables the model to better capture spatial relationships and constraints between them. The model can identify potential collisions and navigate around them. This is valuable in constrained environments where agents must avoid collisions while planning their trajectories. Predicting the closest distance between agents can also help the model understand interactions such as yielding or merging, leading to more accurate forecasts in these complex scenarios.
\subsubsection{\textbf{Direction of Movement Prediction}}
In this pretext task, the main idea is to learn to predict whether two agents are moving closer or receding from each other based on their past trajectory information. This pretext task focuses on understanding the relative motion between agents and determining the nature of their interaction. By predicting the direction of movement between agents, the model can learn to understand the influence agents exert on each other and gain insights into potential collisions or coordinated movements.

Let the pretext task feature extractor for this task to be denoted by $f_{\text{DM}}$ with the parameters $\boldsymbol{\Theta}_{\text{DM}}$. We consider the ground-truth direction of movement between two agents $i$, which is the target agent, and $j$ to be calculated by the difference in intial and final distances between them. $D^{\text{init}}_{j} = ||p_{\text{target}}^{t=1} - p_j^{t=1}||$, and $D^{\text{final}}_{j} = ||p_{\text{target}}^{t=T_c} - p_j^{t=T_c}||$. The direction of movement can be determined by calculating the difference between the initial and final distances as:
$\text{Dir}_j^{\text{GT}} = D^{\text{final}}_{j} - D^{\text{init}}_{j} $. If $\text{Dir}_j^{\text{GT}} > 0$, then agents are receding. If $\text{Dir}_j^{\text{GT}} < 0$, the agents are moving closer. We want to setup this task as a classification task, and choose three classes with labels $y_j^{\text{DM}}$: 
\begin{equation}\label{chap-6-eq:12} 
\small
y_j^{\text{DM}} = 
\begin{cases}
      0 & \text{if} \quad \text{Dir}_j^{\text{GT}} \geq 2 \\
      1 & \text{if} \quad \text{Dir}_j^{\text{GT}} \leq -2 \\
      2 & \text{otherwise}
    \end{cases}     
\normalsize
\end{equation}
The auxiliary direction of motion feature predictor takes the following inputs: the agent-to-agent interaction feature from the standard motion forecasting architecture \cite{Lanegcn} as described in \cref{sec:chap-6-standard model} 
represented by $\tilde{\boldsymbol{v}}_{\text{target}}$ and $\tilde{\boldsymbol{v}}_{j}$, and the distance between them at $t_p=0$ given by $d^0_{\text{target}:j}$. The output is a predicted direction of movement between the target agent and $j$. 
 \begin{equation}\label{chap-6-eq:13} 
\small
{d}_j^{\text{DM}} = \text{Softmax} (f_{\text{DM}}(\tilde{\boldsymbol{v}}_{\text{target}} -  \tilde{\boldsymbol{v}}_j, d^0_{\text{target}:j}))
\normalsize
\end{equation}

The objective function $\boldsymbol{\mathcal{L}}_{\text{DM}}$ for guiding the direction of motion prediction is formulated using cross-entropy loss as shown in \cref{chap-6-eq:14}. $k_\text{target}$ is the number of interacting agents with respect to the target agents as obtained from \cref{sec:chap-6-dataset curation}.
\begin{equation}\label{chap-6-eq:14} 
\small
\boldsymbol{\mathcal{L}}_{\text{DM}}(\boldsymbol{\Theta}_{\text{DM}}) = \frac{1}{k_\text{target}} \sum_{j=1}^{k_\text{target}}{L_\text{cross-entropy}}({d}_j^{\text{DM}}, y_j^{\text{DM}})
\normalsize
\end{equation}

\textit{Benefit to final forecasting performance:} By predicting the direction of movement between agents, the model learns to represent and capture the relative motion between them more effectively. If the agents are predicted to move closer, the model can anticipate a possible collision and adjust the motion forecasts to avoid it. Predicting the direction of movement can help the model identify coordinated movements between agents, such as merging. This is also useful in scenarios when vehicles are accelerating or decelerating, where ego can adjust their trajectories to move safely. In a multi-lane traffic scenario, vehicles may need to change lanes to navigate around slower vehicles or prepare for an upcoming exit. Predicting the direction of movement between vehicles in adjacent lanes can help the model identify when a vehicle is actively changing lanes or maintaining its position. 
\subsubsection{\textbf{Type of Interaction Prediction}}
\label{sec:chap-6-typeofinteraction}
In this pretext task, the main idea is to enhance interaction modeling for trajectory prediction by classifying the types of interactions between agents. This is a classification task that involves predicting among five interaction types $\{$`close-lead', `close-follow', `left-turn-lead', `left-turn-follow', `weak'$\}$, based on the past trajectory information of agents. By classifying the type of interaction, the model can better capture the nuanced dynamics of agent behavior, cooperation and collision, which can lead to more accurate motion forecasts.
Let the pretext task feature extractor for this task to be denoted by $f_{\text{TI}}$ with the parameters $\boldsymbol{\Theta}_{\text{TI}}$. We first describe how we create the pseudo ground-truth labels for this setup, influenced by M2I \cite{M2I}. We consider the type of interaction between two agents $i$, which is the target agent, and $j$. We want to setup this task as a classification task with the pseudo-label for agent $j$ represented as $y_j^{\text{TI}}$. We first check the target agent intention (obtained from \cref{sec:chap-6-dataset curation}. If the target agent intention is straight, we classify the $j$ agents not in the same lane as `weak'. If the target agent intention is `left-turn' or `left-turn-waiting', we label the agents $j$ in the right neighboring lane as `weak'. If the target agent intention is `right-turn' or `right-turn-waiting', we label the agents $j$ in the left neighboring lane as `weak'. \par For the strong interactions, we first compute the closest spatial distance the target agent and agent $j$:
\begin{equation}\label{chap-6-eq:15} 
\small
d_I = \min_{t_1=1}^{T_c} \min_{t_2=1}^{T_c}||p_{\text{target}}^{t_1} - p_j^{t_2}||
\normalsize
\end{equation}
If $d_I > \epsilon_d$, where $\epsilon$ is a user defined threshold, the agents never get close to each other and thus we label the relation $y_j^{\text{TI}}$ as `weak'. Otherwise, we
obtain the time step from each agent at which they reach the closest spatial distance, such that:
\begin{equation}\label{chap-6-eq:16} 
\small
t_1 = \arg \min_{t_1=1}^{T_c} \min_{t_2=1}^{T_c}||p_{\text{target}}^{t_1} - p_j^{t_2}|| 
\normalsize
\end{equation}
\begin{equation}\label{chap-6-eq:17} 
\small
t_2 = \arg \min_{t_2=1}^{T_c} \min_{t_1=1}^{T_c}||p_{\text{target}}^{t_1} - p_j^{t_2}||
\normalsize
\end{equation}
When $t1 > t2$, we define that target agent \textit{follows} agent $j$, as it takes longer for target agent to reach the interaction point. The label $y_j^{\text{TI}}$ for agent $j$ is `left-turn-lead' if the target intention is $\{$`left', `left-turn-waiting'$\}$, and `close-lead' otherwise.
Similarly,  if $t1 < t2$, we define that target agent \textit{leads} agent $j$. The label $y_j^{\text{TI}}$ for agent $j$ is `left-turn-follow' if the target intention is $\{$`left', `left-turn-waiting'$\}$, and `close-follow' otherwise. 

The auxiliary direction of motion feature predictor takes the following inputs: the agent-to-agent interaction feature from the standard motion forecasting architecture \cite{Lanegcn} as described in \cref{sec:chap-6-standard model} 
represented by $\tilde{\boldsymbol{v}}_{\text{target}}$ and $\tilde{\boldsymbol{v}}_{j}$, and the distance between them at $t_p=0$ given by $d^0_{\text{target}:j}$. The output is a predicted type of interaction relation between the target agent and $j$. 
 \begin{equation}\label{chap-6-eq:18} 
\small
{d}_j^{\text{TI}} = \text{Softmax} (f_{\text{TI}}(\tilde{\boldsymbol{v}}_{\text{target}} -  \tilde{\boldsymbol{v}}_j, d^0_{\text{target}:j}))
\normalsize
\end{equation}

The objective function $\boldsymbol{\mathcal{L}}_{\text{TI}}$ for guiding the type of interaction relation prediction is formulated using cross-entropy loss as shown in \cref{chap-6-eq:19}. $k_\text{target}$ is the number of interacting agents with respect to the target agents as obtained from \cref{sec:chap-6-dataset curation}.
\begin{equation}\label{chap-6-eq:19} 
\small
\boldsymbol{\mathcal{L}}_{\text{TI}}(\boldsymbol{\Theta}_{\text{TI}}) = \frac{1}{k_\text{target}} \sum_{j=1}^{k_\text{target}}{L_\text{cross-entropy}}({d}_j^{\text{TI}}, y_j^{\text{TI}})
\normalsize
\end{equation}

\textit{Benefit to final forecasting performance:} Knowing the type of interaction can aid the model in making more informed decisions when predicting trajectories. For example, if the model classifies an interaction as `close-follow', it can adjust the following agent's trajectory to maintain a safe distance from the leading agent. At intersections, vehicles may engage in different types of interactions, such as taking a left turn in front of another vehicle or following closely behind a vehicle making a left turn. In multi-lane traffic scenarios, vehicles may engage in different types of interactions, such as leading or following another vehicle during a lane change or overtaking maneuver. By predicting the type of interaction (e.g., `close-lead', `close-follow'), the model can better understand the dynamics of lane changing and overtaking. This leads to more accurate motion forecasts that account for the specific interaction types.
\subsection{Training Scheme}
\label{sec:chap-6- training scheme}
We train both the pretext losses and the main task in a simultaneous manner. A crucial aspect of our approach is the design of loss functions that specifically aid in the modeling of interactions between agents. To achieve this objective, we restrict the propagation of pretext loss gradients solely to the agent-to-agent interaction module. In practice, this can be accomplished through the use of a stop-gradient operation. We only back-propagate the loss through the target agent of interest and the other agents that interact with it, given a particular scene. Furthermore, since we predict $K$ modes, the pretext task predictions for interactions must reflect this. Therefore, we predict $K$ modes for each of the following functions: $f_{RG}$, $f_{CD}$, $f_{DM}$, and $f_{TI}$. For each target agent, we select the mode that produces the minimum loss with respect to the main forecasting task. 

\begin{table*}[t]
\small
\centering
\resizebox{0.8\textwidth}{!}{%
\begin{tabular}{@{}r|cc|ccc|c@{}}
\toprule
\textbf{Model}                &  \textbf{$\textrm{A2A}$}  &
\textbf{$\textrm{Pretext Task}$}  &
\textbf{$\textrm{minFDE}_1 \downarrow$} & \textbf{$\textrm{minFDE}_6 \downarrow$} & \textbf{$\textrm{MR}_6 \downarrow$} &
\textbf{$\textrm{Improvement}_{\text{minFDE}_6} \uparrow$} \\ \midrule
Without A2A                       & \xmark                       & \xmark                        & 4.006 & 1.516 & 19.879 & -                    \\ 
Without A2A With Pretext                      & \xmark                       & \cmark (All)                     & 3.999 & 1.512 & 19.441 & - \\ 
Baseline                       & \cmark                       & \xmark                        & 3.814 & 1.351 & 16.686 & - \\ 
\midrule 
Ours & \cmark & Range-gap & 3.253 & 1.305 & 15.382 & +3.4\% \\ 
Ours & \cmark & Closest-distance & \textbf{3.230} & 1.295 & 15.209 &    +4.1\% \\ 
Ours & \cmark & Direction of Movement & 3.284 & \textbf{1.279} & \textbf{14.866} &    +5.3\% \\ 
Ours & \cmark & Type Of Interaction & 3.340 & 1.306 & 15.539 &    +3.3\% \\ 
\bottomrule
\end{tabular}%
}
\vspace{0.05in}
\caption{Motion forecasting performance for curated interactive validation set with and without our proposed pretext tasks}
\label{tab:ablation}
\vspace{-0.1in}
\end{table*}

\section{Experiments}
\label{sec:chap-6-experiments}
To train and evaluate the model using pretext tasks, we first introduce the dataset used. Then we describe the models used to evaluate forecasting performance, and illustrate the situations where pretext tasks can enhance predictions.
\subsection{Dataset}
The Argoverse v1.1 \cite{Argoverse} platform offers a comprehensive dataset designed for the training and evaluation of models. The main objective here is the prediction of 3 seconds of future movements, leveraging 2 seconds of past observations. There is a clear division between training and validation sets, avoiding any geographical overlap. In particular, each sequence presents the positions of every actor in a scene, annotated at a frequency of 10Hz, for the past 2 seconds. Each sequence identifies an `agent' of interest whose future 3-second movement is considered for assessment. Both training and validation sets also indicate the future positions of all actors within a 3-second time frame, labeled at 10Hz. 
The dataset provides HD map information for all sequences labeled in Miami and Pittsburgh. The original training set consists of 205,942 sequences, while the validation set has 39,472 sequences. We curate and augment the Argoverse v1.1 using \cref{chap-6-algorithm1} with interaction information and pretext task pseudo labels, as described in \cref{sec:chap-6-dataset curation} and \cref{sec:chap-6-proposed pretext task}. In terms of data statistics, this results in 93,200 unique training sequences and 18592 unique validation sequences. In terms of size, the Waymo Interactive Split \cite{WOMD} and our dataset are comparable. While Waymo Interactive Split assesses on 20,000 scenes with 2 agents, our validation set tests 18,502 scenes, encompassing 47,400 agents. Comparatively, the Waymo Interactive Split, features only two interacting agents per scene, whereas our samples may contain two or more interacting agents.
\subsection{Metrics}
\label{sec:chap-6-proposed metric}
We follow the Argoverse \cite{Argoverse} benchmark and use the
following metrics for evaluation: minimum final displacement error(min-FDE), and miss rate (MR). Where the choice of K is required,
to define the top K trajectories to be used for evaluation of a metric (for example min-FDE (K = k) on Argoverse), we use k=6. 

\begin{table*}[ht]
\small
\centering
\resizebox{0.8\textwidth}{!}{%
\begin{tabular}{@{}r|cc|cccc@{}}
\toprule
\textbf{Model}                &  \textbf{$\textrm{A2A}$}  &
\textbf{$\textrm{Pretext Task}$}  &  \multicolumn{2}{c}{\textbf{$\textrm{i-minFDE}_6 \downarrow$}} & \textbf{$\textrm{ni-min-FDE}_6 \downarrow$} &
\textbf{$\textrm{CAM}_6 \downarrow$} \\ 
 &&& All-Interactive & Strong-Interactive  && \\
\midrule
Baseline                       & \cmark                       & \xmark &  1.279 & 1.320 & 1.459 & 1.472  \\ 
\midrule 
Ours & \cmark & Range-gap & 1.202 (+6.0\%) & 1.242 (+5.9\%) &  1.468 & 1.394 (+5.3\%) \\ 
Ours & \cmark & Closest-distance & 1.192 (+6.8\%)  & 1.235 (+6.4\%) & 1.446 & \textbf{1.346} (+8.6\%) \\ 
Ours & \cmark & Direction of Movement & \textbf{1.175} (+8.1\%) & \textbf{1.216} (+7.9\%) & 1.441 & 1.375 (+6.6\%) \\ 
Ours & \cmark & Type Of Interaction & 1.183 (+7.5\%) & 1.226 (+7.1\%) & 1.455 & 1.376 (+6.5\%) \\ 
\bottomrule
\end{tabular}%
}
\vspace{0.05in}
\caption{Motion forecasting performance for interactive agents curated interactive validation set with our proposed pretext metrics}
\label{tab:proposedmetrics}
\vspace{-0.2in}
\end{table*}

We also propose three additional metrics to assess the quality of interaction prediction. The first one, termed as interactive min-FDE (i-min-FDE), is calculated for all agents interacting with the agent of interest. The i-min-FDE is examined in two scenarios: one including all interacting agents, and the other involving only strongly interacting agents, explicitly excluding any weak interactions as delineated in \cref{sec:chap-6-typeofinteraction}. The second one is termed as non-interactive-min-FDE (ni-min-FDE) which is calculated for all non-interactive sequences. The motivation here is that since pretext tasks are proposed to improve interaction modeling, non-interactive sequences should have similar performance as baseline. The third metric called Collision Avoidance Measure (CAM) is designed to evaluate the performance of predictive models in terms of their ability to forecast and avoid potential collision scenarios between multiple agents in a given environment. The CAM is calculated by iterating over all time steps and pairwise combinations of agents. For each pair of agents $i$ and $j$ at time $t$, we calculate the Euclidean distance between their predicted positions and their actual positions. If predicted distance is less than a distance-threshold (indicating a predicted collision or near-collision scenario) and ground-truth is greater than or equal to the distance-threshold (indicating that no collision or near-collision actually occurred), we increment a counter.
The final value of CAM is the total number of these predicted collisions or near-collision scenarios across all time steps and pairs of agents, divided by the total number of scenes. $\text{CAM} = \frac{1}{s_\text{total}} \sum_{t=1}^{T} \sum_{i=1}^{N-1} \sum_{j=i+1}^{N} \mathcal{I}\{||\mathcal{Y}_{i,t} - \mathcal{Y}_{j,t}|| < d_\text{CAM} \cap ||\mathcal{Y}^{GT}_{i,t} - \mathcal{Y}^{GT}_{j,t}|| >= d_\text{CAM}\}$.
Here, the function $\mathcal{I}(.)$ is an indicator function that equals 1 if the condition in parentheses is true and 0 otherwise, and $s_\text{total}$ is the total number of evaluation sequences. Thus, a lower CAM indicates a better predictive model in terms of its ability to avoid potential collisions.
\subsection{Implementation Details} 
We discuss data pre-processing, training details and base model architecture in this section.

\subsubsection{Data pre-processing:} In order to achieve normalize data, the coordinate system corresponding to each sequence is subjected to translation and rotation. The objective is to reposition the origin at the present position of the acting agent (defined as 'target agent') when $t = 0$, while aligning the positive x-axis with the agent's current direction. This direction is determined based on the orientation of the agent between the location at $t = -1$ and the location at $t = 0$.

\subsubsection{Training:} The model input includes all actors and lanes that interact with the target agent, including the agent itself. The model's training is executed on four TITAN-X GPUs with a batch size of 32. The optimizer used was Adam, with an initial learning rate of $1 \times 10^{-3}$. The rate was decayed to $1 \times 10^{-4}$ at 93k steps, taking approximately five hours in total to finish. $\lambda$ in \cref{chap-6-eq:4} is set to $1$.

\subsubsection{Base Model Architecture} 
\label{sec:chap-6-baseline architecture}
For the agent feature extractor, the architecture is similar to Lane-GCN \cite{Lanegcn}\footnote{\url{https://github.com/AutoVision-cloud/SSL-Interactions}}. We use a 1D CNN to process the trajectory input. The output is a temporal feature map, whose element at $t = 0$ is used as the agent feature. The network has three groups/scales of 1D convolutions. Each group consists of two residual blocks, with the stride of the first block as 2. Feature Pyramid Network (FPN) fuses the multi-scale features, and applies another residual block to obtain the output tensor. For all layers, the convolution kernel size is 3 and the number of output channels is 128. Layer normalization and Rectified Linear Unit (ReLU) are used after each convolution. 
The {map feature extractor} has two LaneConv residual blocks  which are the stack of a LaneConv(1) and a linear layer, as well as a shortcut. All layers have 128 feature channels. Layer normalization  and ReLU are used after each LaneConv and linear layer. 
\\ For the map-aware agent feature (M2A) module, the distance threshold is $12$m. The one M2A interaction module has two residual blocks, which consist of a stack of an attention layer and a linear layer, as well as a residual connection. The A2A layer has four such blocks. All layers have 128 output feature channels. The pretext task encoder that takes as input the A2A vector has two residual blocks, which also have 128 dimensional output.
\\ Taking the A2A actor features as input, our trajectory decoder is a multi-modal prediction header that outputs the final motion forecasting. For each agent, it predicts $K=6$ possible future trajectories and confidence scores. The header has two branches, a regression branch to predict the trajectory of each mode and a classification branch to predict the confidence score of each mode.
\subsection{Baselines and Proposed Model}
We train different versions of our base model with and without the interaction component (A2A layer) and the pretext tasks. We compare the performance of these models on our validation set.
We have the following four variants:
\begin{itemize}
    \item Without A2A: Train the base model described in \cref{sec:chap-6-baseline architecture} with the A2A layer, but without the pretext tasks.
    \item Without A2A With Pretext: Train the base model without the A2A layer, but with the pretext task losses added. The A2A layer has a stop gradient for the pretext loss as described in \cref{sec:chap-6- training scheme}.
    \item Baseline: Train the base model without the A2A layer and without the pretext tasks.
    \item Proposed Model: Train the baseline model with both the A2A layer and the proposed pretext tasks. This model also has a stop gradient operation so that the pretext loss only affects the A2A layer.
\end{itemize}

\section{Results}
\label{sec:chap-6-discussion}
\textbf{Ablation:} Our findings reveal several distinct trends, as displayed in \cref{tab:ablation}. This table outlines the metrics of the target agent, as defined by Argoverse, on our specially-selected, interaction-dense validation dataset.

We observe that the model lacking the A2A layer (shown in the first row) under-performs our established baseline (shown in the third row) by a notable 11\%. This underscores the considerable influence of the A2A layer when assessing performance on our interaction-heavy dataset.
\begin{figure*}[t]
  \centering
   \includegraphics[height=12cm]{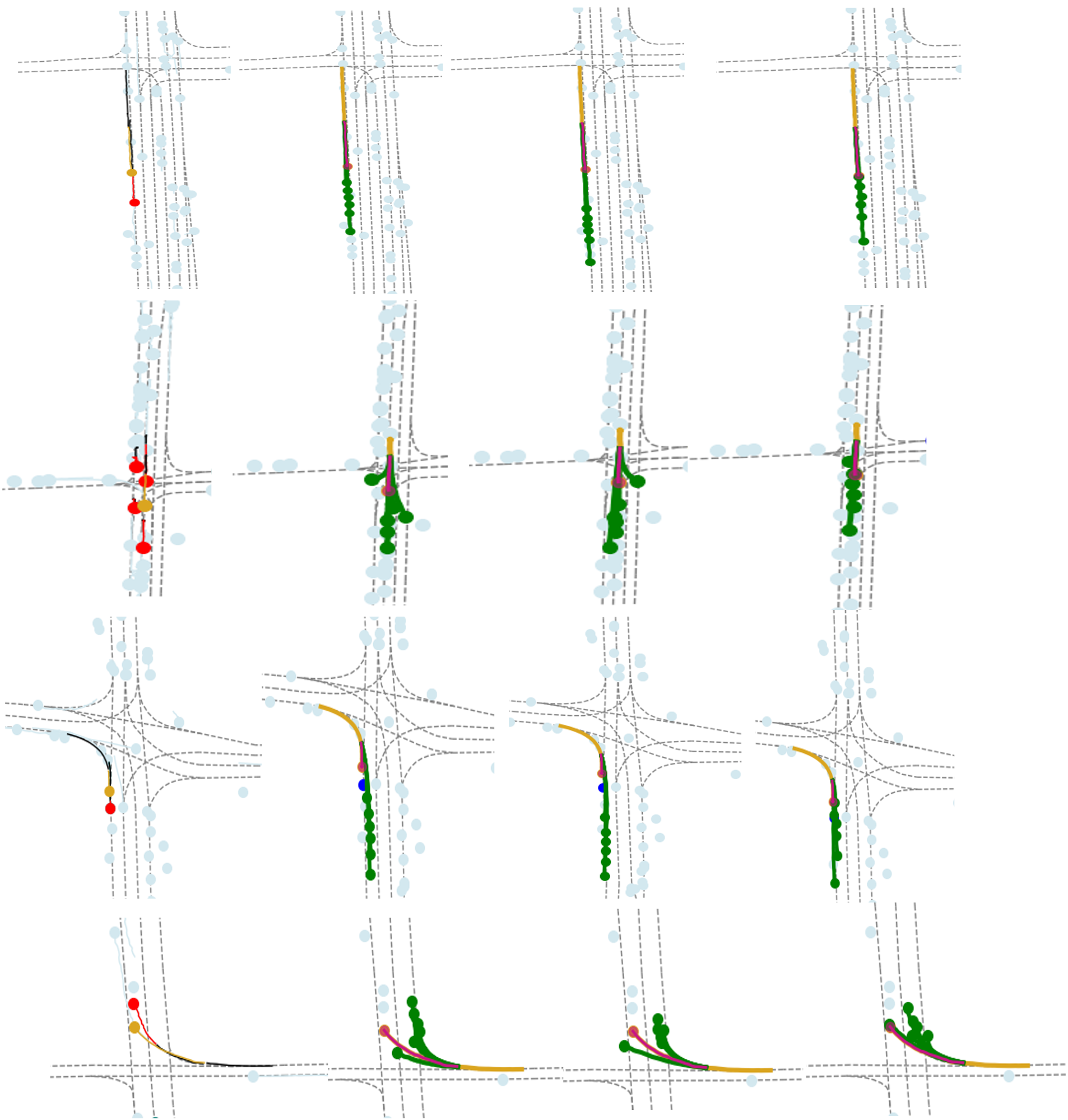}
 \caption{Motion forecasting on curated, interaction-heavy Argoverse \cite{Argoverse} validation. We present four challenging scenarios for analysis. The first column depicts the scene featuring the target agent and the curated interactive agents. The second column contains predictions made by the baseline model \cite{Lanegcn}. The third column displays the predictions of our proposed model when the connections to the interactive agents are disconnected. The fourth and final column features predictions from our model when regularized with the pretext loss. The baseline model fails to accurately forecast any of the scenarios. The first row illustrates a case where the predicted range-gap accurately anticipates at least one future trajectory that evades collision with the forward vehicle. The second row presents a congested situation in which forecasting the closest-distance with interacting agents leads to a future trajectory devoid of collision with the vehicle in front. The third row depicts a similar situation, but in the context of predicting direction of movement - both the baseline and the model without interacting agent information result in a collision course with the agent ahead, whereas our model proposes a trajectory closely aligned with the ground truth, avoiding collision.The final row presents a turning scenario wherein the type of interaction is predicted. Our proposed model forecasts a `close-follow' situation with the interacting vehicle and identifies a potential future trajectory that is most closely aligned with the ground truth, while the baseline and the model without knowledge of interacting agents fail in terms of the $\text{MR}_6$ metric. Please refer to \cref{sec:chap-6-proposed pretext task} for details.}
   \label{chap-6-qual}
   \vspace{-0.4cm}
\end{figure*}
The second model, which is without the A2A layer but incorporates pretext tasks (calculated as the cumulative loss from all proposed pretext tasks), mirrors the performance of the first model closely. This suggests that the pretext tasks do not impact the M2A layer, the map, or the agent encoder, due to the application of the stop gradient operation. Instead, they only influence the A2A layer. This finding is pivotal in validating our proposition that the pretext tasks assist in interaction modeling.

\textbf{Comparison with state-of-the-art:} Across all proposed pretext tasks, we also noticed a significant improvement in the model’s performance w.r.t state-of-the-art baseline — ranging from 3.3\% to 5.3\% across the $\text{minFDE}_6$ metric. We centered our attention on min-FDE in this table because of its importance in making precise long-term predictions, which are vital for safety. The most effective predictors for this dataset were direction of movement and closest distance. The range-gap prediction and the type of interaction prediction under-perform the above mentioned tasks. In examining the reasons for this disparity, we hypothesize that the structure of the range-gap prediction task, which is a regression task, could have contributed to its lesser performance. Regression tasks are typically more challenging to learn compared to classification tasks due to their demand for more complex learning mechanisms. Furthermore, the task of type of interaction prediction presented two significant difficulties. Firstly, the task was heavily imbalanced, and secondly, it did not offer precise information about the trajectory endpoints. The lack of fine-grained information in this task could have inhibited the model's ability to make accurate long-term predictions. These observations underscore the need to carefully select and design pretext tasks when building models for interaction-heavy datasets.

\textbf{Evaluation on Proposed Metrics:} A consistent pattern can be observed in \cref{tab:proposedmetrics}, where we evaluate our proposed metrics for both the baseline and the suggested pretext tasks. We assess the performance using the metrics $\text{i-minFDE}_6$ and $\text{CAM}_6$, not only for the target agent but also for all interacting agents. This comprehensive evaluation is crucial as it is not only the predictions for the primary agent that matter, but also those for the surrounding agents in an interactive scenario. Our findings indicate that there are more substantial improvements (upto 8.1\% for all interacting agents, 7.9\% for only strongly interacting agents, 8.6\% for collision awareness) over the baseline when using these proposed metrics. This suggests that the final positions of the interacting agents are predicted more accurately than with the baseline, and there's improved collision awareness. These metrics provide a more detailed understanding of the model's performance in predicting interactive scenarios. 

In addition to the interactive dataset, we maintain a set of non-interactive validation data. Upon evaluating this data using the $\text{ni-minFDE}_6$ metric, we observe that the baseline model's performance is nearly on par with the models employing pretext tasks. This observation is interesting — it indicates that the improvements resulting from the inclusion of pretext losses are primarily manifested in datasets with heavy interaction. Specifically, the pretext tasks do not notably enhance performance on non-interactive data. This finding aligns with our objective to demonstrate that pretext tasks are particularly beneficial in modeling interactions.

\textbf{Qualitative Results:} Illustrations of improvements provided by the pretext task models are shown in \cref{chap-6-qual}. Each illustration portrays a unique urban driving scenario, and details how the introduction of the pretext task model enhances the predictive capabilities. In these diagrams, we compare the ground-truth with the baseline model's prediction, the model's prediction when interactions with other agents have been disconnected, and the proposed model's prediction. These illustrations serve to underscore the effectiveness of our pretext task models in a real-world context.

\section{Conclusion}
\label{sec:chap6-conclusion}
In this work, we improve motion forecasting with a particular focus on interactive scenarios. Our first contribution is a simple yet effective approach for interaction modeling. We introduce four interaction-specific pretext tasks, which incorporate domain knowledge and are learned concurrently with the main task. This methodology enables the pretext task to effectively capture the dependencies between agents' movements. Our second contribution is a novel method for curating datasets, which allows for the explicit labeling of interacting pairs. This approach is crucial for the training of interaction-specific pretext tasks and the generation of necessary pseudo-labels. The third contribution is the development of new metrics designed to offer a more accurate evaluation of performance in scenarios characterized by a high degree of interaction. These metrics provide a detailed and nuanced understanding of the model's performance. Finally, we demonstrate that our proposed methodology significantly outperforms the baseline in both quantitative measurements and qualitative assessments. In summary, our work advances the understanding and application of interaction modeling in motion forecasting, providing useful tools and methods that can be further built upon in future research.

\textit{Limitations:} We acknowledge certain limitations in our current approach. Specifically, performance improvements cannot be simply achieved by aggregating the pretext losses from the proposed tasks. This strategy necessitates meticulous hyper-parameter tuning or a considered selection of task pairs that can mutually enhance each other's performance. We recognize this as an area of potential exploration and refinement for future research. Additionally, we acknowledge the need to extend our research to cover more diverse scenarios, including interactions involving pedestrians and cyclists. Lastly, it would be beneficial to validate the effectiveness of our proposed pretext tasks across a broader array of datasets. 


\bibliographystyle{IEEEtran}
\bibliography{root}

\begin{thebibliography}{10}
\providecommand{\url}[1]{#1}
\csname url@rmstyle\endcsname
\providecommand{\newblock}{\relax}
\providecommand{\bibinfo}[2]{#2}
\providecommand\BIBentrySTDinterwordspacing{\spaceskip=0pt\relax}
\providecommand\BIBentryALTinterwordstretchfactor{4}
\providecommand\BIBentryALTinterwordspacing{\spaceskip=\fontdimen2\font plus
\BIBentryALTinterwordstretchfactor\fontdimen3\font minus \fontdimen4\font\relax}
\providecommand\BIBforeignlanguage[2]{{%
\expandafter\ifx\csname l@#1\endcsname\relax
\typeout{** WARNING: IEEEtran.bst: No hyphenation pattern has been}%
\typeout{** loaded for the language `#1'. Using the pattern for}%
\typeout{** the default language instead.}%
\else
\language=\csname l@#1\endcsname
\fi
#2}}

\bibitem{ModelBased}
\BIBentryALTinterwordspacing
W.~Shang, A.~Trott, S.~Zheng, C.~Xiong, and R.~Socher, ``Learning world graphs to accelerate hierarchical reinforcement learning,'' \emph{CoRR}, vol. abs/1907.00664, 2019. [Online]. Available: \url{http://arxiv.org/abs/1907.00664}
\BIBentrySTDinterwordspacing

\bibitem{SocialForces}
\BIBentryALTinterwordspacing
D.~Helbing and P.~Moln{\'{a}}r, ``Social force model for pedestrian dynamics,'' \emph{Physical Review E}, vol.~51, no.~5, pp. 4282--4286, may 1995. [Online]. Available: \url{https://doi.org/10.1103%2Fphysreve.51.4282}
\BIBentrySTDinterwordspacing

\bibitem{Lanegcn}
\BIBentryALTinterwordspacing
M.~Liang, B.~Yang, R.~Hu, Y.~Chen, R.~Liao, S.~Feng, and R.~Urtasun, ``Learning lane graph representations for motion forecasting,'' in \emph{Computer Vision - {ECCV} 2020 - 16th European Conference, Glasgow, UK, August 23-28, 2020, Proceedings, Part {II}}, ser. Lecture Notes in Computer Science, A.~Vedaldi, H.~Bischof, T.~Brox, and J.~Frahm, Eds., vol. 12347.\hskip 1em plus 0.5em minus 0.4em\relax Springer, 2020, pp. 541--556. [Online]. Available: \url{https://doi.org/10.1007/978-3-030-58536-5\_32}
\BIBentrySTDinterwordspacing

\bibitem{M2I}
\BIBentryALTinterwordspacing
Q.~Sun, X.~Huang, J.~Gu, B.~C. Williams, and H.~Zhao, ``{M2I:} from factored marginal trajectory prediction to interactive prediction,'' in \emph{{IEEE/CVF} Conference on Computer Vision and Pattern Recognition, {CVPR} 2022, New Orleans, LA, USA, June 18-24, 2022}.\hskip 1em plus 0.5em minus 0.4em\relax {IEEE}, 2022, pp. 6533--6542. [Online]. Available: \url{https://doi.org/10.1109/CVPR52688.2022.00643}
\BIBentrySTDinterwordspacing

\bibitem{SocialLSTM}
\BIBentryALTinterwordspacing
A.~Alahi, K.~Goel, V.~Ramanathan, A.~Robicquet, L.~Fei{-}Fei, and S.~Savarese, ``Social {LSTM:} human trajectory prediction in crowded spaces,'' in \emph{2016 {IEEE} Conference on Computer Vision and Pattern Recognition, {CVPR} 2016, Las Vegas, NV, USA, June 27-30, 2016}.\hskip 1em plus 0.5em minus 0.4em\relax {IEEE} Computer Society, 2016, pp. 961--971. [Online]. Available: \url{https://doi.org/10.1109/CVPR.2016.110}
\BIBentrySTDinterwordspacing

\bibitem{SocialGAN}
\BIBentryALTinterwordspacing
A.~Gupta, J.~Johnson, L.~Fei{-}Fei, S.~Savarese, and A.~Alahi, ``Social {GAN:} socially acceptable trajectories with generative adversarial networks,'' in \emph{2018 {IEEE} Conference on Computer Vision and Pattern Recognition, {CVPR} 2018, Salt Lake City, UT, USA, June 18-22, 2018}.\hskip 1em plus 0.5em minus 0.4em\relax Computer Vision Foundation / {IEEE} Computer Society, 2018, pp. 2255--2264. [Online]. Available: \url{http://openaccess.thecvf.com/content\_cvpr\_2018/html/Gupta\_Social\_GAN\_Socially\_CVPR\_2018\_paper.html}
\BIBentrySTDinterwordspacing

\bibitem{RN}
\BIBentryALTinterwordspacing
A.~Santoro, D.~Raposo, D.~G.~T. Barrett, M.~Malinowski, R.~Pascanu, P.~W. Battaglia, and T.~Lillicrap, ``A simple neural network module for relational reasoning,'' in \emph{Advances in Neural Information Processing Systems 30: Annual Conference on Neural Information Processing Systems 2017, December 4-9, 2017, Long Beach, CA, {USA}}, I.~Guyon, U.~von Luxburg, S.~Bengio, H.~M. Wallach, R.~Fergus, S.~V.~N. Vishwanathan, and R.~Garnett, Eds., 2017, pp. 4967--4976. [Online]. Available: \url{https://proceedings.neurips.cc/paper/2017/hash/e6acf4b0f69f6f6e60e9a815938aa1ff-Abstract.html}
\BIBentrySTDinterwordspacing

\bibitem{STGCNN}
\BIBentryALTinterwordspacing
A.~A. Mohamed, K.~Qian, M.~Elhoseiny, and C.~G. Claudel, ``Social-stgcnn: {A} social spatio-temporal graph convolutional neural network for human trajectory prediction,'' in \emph{2020 {IEEE/CVF} Conference on Computer Vision and Pattern Recognition, {CVPR} 2020, Seattle, WA, USA, June 13-19, 2020}.\hskip 1em plus 0.5em minus 0.4em\relax Computer Vision Foundation / {IEEE}, 2020, pp. 14\,412--14\,420. [Online]. Available: \url{https://openaccess.thecvf.com/content\_CVPR\_2020/html/Mohamed\_Social-STGCNN\_A\_Social\_Spatio-Temporal\_Graph\_Convolutional\_Neural\_Network\_for\_Human\_CVPR\_2020\_paper.html}
\BIBentrySTDinterwordspacing

\bibitem{SpaGNN}
\BIBentryALTinterwordspacing
S.~Casas, C.~Gulino, R.~Liao, and R.~Urtasun, ``Spagnn: Spatially-aware graph neural networks for relational behavior forecasting from sensor data,'' in \emph{2020 {IEEE} International Conference on Robotics and Automation, {ICRA} 2020, Paris, France, May 31 - August 31, 2020}.\hskip 1em plus 0.5em minus 0.4em\relax {IEEE}, 2020, pp. 9491--9497. [Online]. Available: \url{https://doi.org/10.1109/ICRA40945.2020.9196697}
\BIBentrySTDinterwordspacing

\bibitem{ILVM}
\BIBentryALTinterwordspacing
S.~Casas, C.~Gulino, S.~Suo, K.~Luo, R.~Liao, and R.~Urtasun, ``Implicit latent variable model for scene-consistent motion forecasting,'' in \emph{Computer Vision - {ECCV} 2020 - 16th European Conference, Glasgow, UK, August 23-28, 2020, Proceedings, Part {XXIII}}, ser. Lecture Notes in Computer Science, A.~Vedaldi, H.~Bischof, T.~Brox, and J.~Frahm, Eds., vol. 12368.\hskip 1em plus 0.5em minus 0.4em\relax Springer, 2020, pp. 624--641. [Online]. Available: \url{https://doi.org/10.1007/978-3-030-58592-1\_37}
\BIBentrySTDinterwordspacing

\bibitem{Trajectron++}
\BIBentryALTinterwordspacing
T.~Salzmann, B.~Ivanovic, P.~Chakravarty, and M.~Pavone, ``Trajectron++: Dynamically-feasible trajectory forecasting with heterogeneous data,'' in \emph{Computer Vision - {ECCV} 2020 - 16th European Conference, Glasgow, UK, August 23-28, 2020, Proceedings, Part {XVIII}}, ser. Lecture Notes in Computer Science, A.~Vedaldi, H.~Bischof, T.~Brox, and J.~Frahm, Eds., vol. 12363.\hskip 1em plus 0.5em minus 0.4em\relax Springer, 2020, pp. 683--700. [Online]. Available: \url{https://doi.org/10.1007/978-3-030-58523-5\_40}
\BIBentrySTDinterwordspacing

\bibitem{SocialBiGat}
\BIBentryALTinterwordspacing
V.~Kosaraju, A.~Sadeghian, R.~Mart{\'{\i}}n{-}Mart{\'{\i}}n, I.~D. Reid, H.~Rezatofighi, and S.~Savarese, ``Social-bigat: Multimodal trajectory forecasting using bicycle-gan and graph attention networks,'' in \emph{Advances in Neural Information Processing Systems 32: Annual Conference on Neural Information Processing Systems 2019, NeurIPS 2019, December 8-14, 2019, Vancouver, BC, Canada}, H.~M. Wallach, H.~Larochelle, A.~Beygelzimer, F.~d'Alch{\'{e}}{-}Buc, E.~B. Fox, and R.~Garnett, Eds., 2019, pp. 137--146. [Online]. Available: \url{https://proceedings.neurips.cc/paper/2019/hash/d09bf41544a3365a46c9077ebb5e35c3-Abstract.html}
\BIBentrySTDinterwordspacing

\bibitem{Scenetransformer}
\BIBentryALTinterwordspacing
J.~Ngiam, V.~Vasudevan, B.~Caine, Z.~Zhang, H.~L. Chiang, J.~Ling, R.~Roelofs, A.~Bewley, C.~Liu, A.~Venugopal, D.~J. Weiss, B.~Sapp, Z.~Chen, and J.~Shlens, ``Scene transformer: {A} unified architecture for predicting future trajectories of multiple agents,'' in \emph{The Tenth International Conference on Learning Representations, {ICLR} 2022, Virtual Event, April 25-29, 2022}.\hskip 1em plus 0.5em minus 0.4em\relax OpenReview.net, 2022. [Online]. Available: \url{https://openreview.net/forum?id=Wm3EA5OlHsG}
\BIBentrySTDinterwordspacing

\bibitem{Multiplefutures}
\BIBentryALTinterwordspacing
Y.~C. Tang and R.~Salakhutdinov, ``Multiple futures prediction,'' in \emph{Advances in Neural Information Processing Systems 32: Annual Conference on Neural Information Processing Systems 2019, NeurIPS 2019, December 8-14, 2019, Vancouver, BC, Canada}, H.~M. Wallach, H.~Larochelle, A.~Beygelzimer, F.~d'Alch{\'{e}}{-}Buc, E.~B. Fox, and R.~Garnett, Eds., 2019, pp. 15\,398--15\,408. [Online]. Available: \url{https://proceedings.neurips.cc/paper/2019/hash/86a1fa88adb5c33bd7a68ac2f9f3f96b-Abstract.html}
\BIBentrySTDinterwordspacing

\bibitem{MultiplexAttention}
\BIBentryALTinterwordspacing
F.~Sun, I.~Kauvar, R.~Zhang, J.~Li, M.~J. Kochenderfer, J.~Wu, and N.~Haber, ``Interaction modeling with multiplex attention,'' in \emph{NeurIPS}, 2022. [Online]. Available: \url{http://papers.nips.cc/paper\_files/paper/2022/hash/7e6361a5d73a8fab093dd8453e0b106f-Abstract-Conference.html}
\BIBentrySTDinterwordspacing

\bibitem{TellMeWhy}
\BIBentryALTinterwordspacing
A.~K. Lampinen, N.~A. Roy, I.~Dasgupta, S.~C.~Y. Chan, A.~C. Tam, J.~L. McClelland, C.~Yan, A.~Santoro, N.~C. Rabinowitz, J.~X. Wang, and F.~Hill, ``Tell me why! explanations support learning relational and causal structure,'' in \emph{International Conference on Machine Learning, {ICML} 2022, 17-23 July 2022, Baltimore, Maryland, {USA}}, ser. Proceedings of Machine Learning Research, K.~Chaudhuri, S.~Jegelka, L.~Song, C.~Szepesv{\'{a}}ri, G.~Niu, and S.~Sabato, Eds., vol. 162.\hskip 1em plus 0.5em minus 0.4em\relax {PMLR}, 2022, pp. 11\,868--11\,890. [Online]. Available: \url{https://proceedings.mlr.press/v162/lampinen22a.html}
\BIBentrySTDinterwordspacing

\bibitem{LanguagePretext}
\BIBentryALTinterwordspacing
Y.~Kuo, X.~Huang, A.~Barbu, S.~G. McGill, B.~Katz, J.~J. Leonard, and G.~Rosman, ``Trajectory prediction with linguistic representations,'' in \emph{2022 International Conference on Robotics and Automation, {ICRA} 2022, Philadelphia, PA, USA, May 23-27, 2022}.\hskip 1em plus 0.5em minus 0.4em\relax {IEEE}, 2022, pp. 2868--2875. [Online]. Available: \url{https://doi.org/10.1109/ICRA46639.2022.9811928}
\BIBentrySTDinterwordspacing

\bibitem{DomainPseudoLabels}
\BIBentryALTinterwordspacing
L.~Sun, C.~Tang, Y.~Niu, E.~Sachdeva, C.~Choi, T.~Misu, M.~Tomizuka, and W.~Zhan, ``Domain knowledge driven pseudo labels for interpretable goal-conditioned interactive trajectory prediction,'' in \emph{{IEEE/RSJ} International Conference on Intelligent Robots and Systems, {IROS} 2022, Kyoto, Japan, October 23-27, 2022}.\hskip 1em plus 0.5em minus 0.4em\relax {IEEE}, 2022, pp. 13\,034--13\,041. [Online]. Available: \url{https://doi.org/10.1109/IROS47612.2022.9982147}
\BIBentrySTDinterwordspacing

\bibitem{SocialSSL}
\BIBentryALTinterwordspacing
L.~Tsao, Y.~Wang, H.~Lin, H.~Shuai, L.~Wong, and W.~Cheng, ``Social-ssl: Self-supervised cross-sequence representation learning based on transformers for multi-agent trajectory prediction,'' in \emph{Computer Vision - {ECCV} 2022 - 17th European Conference, Tel Aviv, Israel, October 23-27, 2022, Proceedings, Part {XXII}}, ser. Lecture Notes in Computer Science, S.~Avidan, G.~J. Brostow, M.~Ciss{\'{e}}, G.~M. Farinella, and T.~Hassner, Eds., vol. 13682.\hskip 1em plus 0.5em minus 0.4em\relax Springer, 2022, pp. 234--250. [Online]. Available: \url{https://doi.org/10.1007/978-3-031-20047-2\_14}
\BIBentrySTDinterwordspacing

\bibitem{Argoverse}
\BIBentryALTinterwordspacing
M.~Chang, J.~Lambert, P.~Sangkloy, J.~Singh, S.~Bak, A.~Hartnett, D.~Wang, P.~Carr, S.~Lucey, D.~Ramanan, and J.~Hays, ``Argoverse: 3d tracking and forecasting with rich maps,'' in \emph{{IEEE} Conference on Computer Vision and Pattern Recognition, {CVPR} 2019, Long Beach, CA, USA, June 16-20, 2019}.\hskip 1em plus 0.5em minus 0.4em\relax Computer Vision Foundation / {IEEE}, 2019, pp. 8748--8757. [Online]. Available: \url{http://openaccess.thecvf.com/content\_CVPR\_2019/html/Chang\_Argoverse\_3D\_Tracking\_and\_Forecasting\_With\_Rich\_Maps\_CVPR\_2019\_paper.html}
\BIBentrySTDinterwordspacing

\bibitem{WOMD}
\BIBentryALTinterwordspacing
S.~Ettinger, S.~Cheng, B.~Caine, C.~Liu, H.~Zhao, S.~Pradhan, Y.~Chai, B.~Sapp, C.~R. Qi, Y.~Zhou, Z.~Yang, A.~Chouard, P.~Sun, J.~Ngiam, V.~Vasudevan, A.~McCauley, J.~Shlens, and D.~Anguelov, ``Large scale interactive motion forecasting for autonomous driving : The waymo open motion dataset,'' in \emph{2021 {IEEE/CVF} International Conference on Computer Vision, {ICCV} 2021, Montreal, QC, Canada, October 10-17, 2021}.\hskip 1em plus 0.5em minus 0.4em\relax {IEEE}, 2021, pp. 9690--9699. [Online]. Available: \url{https://doi.org/10.1109/ICCV48922.2021.00957}
\BIBentrySTDinterwordspacing

\bibitem{PretextTaskasRegularization}
\BIBentryALTinterwordspacing
L.~Liebel and M.~K{\"{o}}rner, ``Auxiliary tasks in multi-task learning,'' \emph{CoRR}, vol. abs/1805.06334, 2018. [Online]. Available: \url{http://arxiv.org/abs/1805.06334}
\BIBentrySTDinterwordspacing

\end{thebibliography}

\section*{Appendix}
\label{appendix}

\subsection{Standard Interaction Modeling}
\label{sec:chap-6-standard model}
The encoder for the input motion of the agents is parameterized by $f_{\text{enc}}$, and the encoding can be represented by \cref{chap-6-eq:1}. To incorporate contextual information such as lane semantics, a graph convolutional neural network (GCN) parameterized by $f_{\text{map}}$, is employed to operate on a graph representation of the scene depicted in \cref{chap-6-eq:2} and called a M2A layer. Finally, to model agent-to-agent interactions, the representation of nearby agents is aggregated using a graph convolutional neural network (GCN) parameterized by $f_{\text{interact}}$. This is called an A2A layer. The graph representation is constructed such that agents are represented as nodes in the graph, and edges are established between agents if they are within a certain distance of each other. Specifically, the neighboring set of node $i$ is represented as $\mathcal{N}_e$, and the distance between agents $i$ and $j$ is given by $d_{ij}$. The input to the GCN includes the agent $i$'s embedding $\boldsymbol{\hat{v}}_i$, the contextual embedding obtained for each neighboring agent $\{\boldsymbol{\hat{v}}_j | j \in \mathcal{N}_e\}$, and $d_{ij}$. During the GCN operation, the interaction information from neighboring agent states is \textit{implicitly} incorporated by performing message passing between nodes in the graph, as depicted in \cref{chap-6-eq:3}. By incorporating the temporal representation of nearby agents and their relative distances into the final agent state $\boldsymbol{\tilde{v}}_i$ to be used for future prediction, agent-to-agent interactions are modeled within the data-driven motion forecasting framework. 
\begin{equation}\label{chap-6-eq:1} 
\small
\hat{\boldsymbol{s}}_i = f_{\text{enc}}(\mathcal{S}_i) 
\normalsize
\end{equation}
\begin{equation}\label{chap-6-eq:2} 
\small
\hat{\boldsymbol{v}}_i = f_{\text{map}}(\hat{\boldsymbol{s}}_i, \boldsymbol{h}_0) 
\normalsize
\end{equation}
\begin{equation}\label{chap-6-eq:3}
\small
\boldsymbol{\tilde{v}}_i =  \sum_{j \in \mathcal{N}_e}{f_{\text{interact}}(\boldsymbol{\hat{v}}_i, \boldsymbol{\hat{v}}_j, d_{ij})}
\normalsize
\end{equation}
\subsection{Pretext Task Regularization}
\label{sec:chap6-pretext}
Pretext task regularization is a technique that involves training a model on a auxiliary task, called a pretext task, in order to improve the model's performance on the main task. The idea is to leverage the structure of the pretext task to encourage the model to learn representations that are transferable to the main task of interest \cite{PretextTaskasRegularization}. A pretext task is designed using domain-knowledge to capture some inherent structure or pattern in the data. The most effective pretext tasks exploit the underlying structure in such a way that the learned representations are meaningful and generalizable. This does not add to model complexity, as the pretext tasks parameters are discarded at inference time.

The existing literature for trajectory prediction predominantly focuses on supervised learning scenarios where the learner has access to a large, annotated dataset that can be revisited multiple times to learn the optimal feature extractor $f_{\theta}$. However, in numerous real-world scenarios, we often have access to vast amounts of unlabeled data, which can potentially make the learning process more efficient and cost-effective. In this context, our objective is to investigate if designing pretext tasks that emphasize the semantics of inter-object interactions can lead to more meaningful representations for agents, ultimately enhancing the model's ability to make accurate future predictions.
\par  To incorporate a pretext task into the learning process, we train a model on both the main task and the pretext task simultaneously. Let the model parameters be denoted by $\boldsymbol{\Theta}$. Let the main task loss function be $L_{\text{main}}(\boldsymbol{\Theta})$, and the pretext task loss function by $L_{\text{pretext}}(\boldsymbol{\Theta})$. The overall loss function for the combined learning problem can be expressed as \cref{chap-6-eq:4}. $\lambda$ is a regularization hyper-parameter that controls the balance between the main task and the pretext task. The model's parameters are updated using gradient-based optimization. 
\begin{equation}\label{chap-6-eq:4}
\small
L_{\text{total}}(\boldsymbol{\Theta}) = L_{\text{main}}(\boldsymbol{\Theta}) + \lambda * L_{\text{pretext}}(\boldsymbol{\Theta})
\normalsize
\end{equation}
In summary, pretext tasks provide a powerful framework for leveraging the structure of unlabeled data to learn useful representations that can be transferred to the main task. 
\subsection{Design Considerations for Interaction Modeling}
\label{sec:chap-6-design}
\quad \textit{(A) Marginal Prediction:} 
\label{marginal-predictions}
In general, a trajectory prediction model learns to model the distribution $p({\boldsymbol{\mathcal{Y}}}|\mathcal{M}, \boldsymbol{h}_0)$. In the case of multiple agents, trajectory prediction is often evaluated independently for each agent. This means that accurately predicting the marginal distribution of vehicle trajectories is sufficient for achieving good results on benchmark tasks. Although the agent-to-agent (A2A) layer \textit{implicitly} models interactions within the framework, the training parameters may focus solely on optimizing the final forecasting loss. In \cref{sec:chap-6-standard model}, we consider a representative motion forecasting model of this type. This model captures interactions between agents implicitly, using message passing via graph convolutions to model $p(\mathcal{Y}_i|\mathcal{M}, \boldsymbol{h}_0)$. However, models that focus solely on predicting marginal distributions may generate joint behaviors that are infeasible or unrealistic.
\par \textit{(B) Full Joint Prediction:}  To properly evaluate and assess such interactive behaviors, it is necessary to predict the joint distribution of the future trajectories of all interacting agents. Directly modeling the joint distribution $p(\mathcal{Y}_1, \mathcal{Y}_2, ..., \mathcal{Y}_N |\mathcal{M}, \boldsymbol{h}_0)$ however grows exponentially with the number of agents. This presents a significant challenge for models that aim to accurately predict joint behaviors for complex interactions between multiple agents.
\par \textit{(C) SSL-Interactions:}
We propose to use pseudo-labeled interacting pairs to train the interactions \textit{explicitly} via pretext tasks. A pretext task is defined as $T$. Each pretext task $T$ is associated with a subset of agents $A_i \in {\{\mathcal{Y}_1, \mathcal{Y}_2, ..., \mathcal{Y}_{k_i}}\}$, where $k_i < N$ is the number of agents involved in task $T$ for agent $i$. Under the assumption that the pretext tasks capture important conditional independence properties among the agents, we can factorize the joint probability distribution $p(\mathcal{Y}_1, \mathcal{Y}_2, ..., \mathcal{Y}_N |\mathcal{M}, \boldsymbol{h}_0)$ into a product of a marginal distribution $p_0(\mathcal{Y}_i|\mathcal{M}, \boldsymbol{h}_0)$ and pretext task-specific conditional distributions $P(T|A_i)$. This is shown in \cref{chap-6-eq:5}. 
\begin{equation}\label{chap-6-eq:5} 
\small
p(\mathcal{Y}_1, \mathcal{Y}_2, ..., \mathcal{Y}_N |\mathcal{M}, \boldsymbol{h}_0) = \prod_{i=1}^N{p_0(\mathcal{Y}_i|\mathcal{M}, \boldsymbol{h}_0)} \prod_{i=1}^N{P(T | A_i)}
\normalsize
\end{equation}
$p_0(\mathcal{Y}_i|\mathcal{M}, \boldsymbol{h}_0)$ is a distribution that captures the marginal dependencies among the agents as described in \cref{marginal-predictions}(A), and drives $\mathcal{L}_{main}$ in \cref{architecture}. $P(T | A_i)$ is the conditional probability distribution of pretext task $T$ given the states of the agents in $A_i$, which captures interaction information between them \textit{explicitly}, and drives $\mathcal{L}_{pretext}$ in \cref{architecture}.

The factorization in \cref{chap-6-eq:5} reduces the complexity of the joint distribution by decomposing it into simpler components. To model each component, one can employ maximum likelihood estimation. For the our proposed pretext tasks \cref{sec:chap-6-proposed pretext task}, the negative log-likelihood loss for the pretext tasks corresponds to a Mean Squared Error (MSE) loss for regression setting when we incorporate the Gaussian likelihood assumption, and cross-entropy loss for the classification setting.  
\par Identifying the optimal decomposition and pretext tasks for a given problem can be a challenging task, and may require domain-specific knowledge. In this chapter, we propose both pretext tasks tailored for interactive trajectory prediction, and a method to pseudo-label the optimal subset of interacting agents $A_i$. 
\section{Detail of Algorithm for Labeling Interactions}
\par We propose a simple but effective approach to identify interacting pairs of trajectories based on the spatial distance between the agents and their intents. The method consists of three main steps, as described below. We do not label trajectories if they are too short.
\begin{itemize}
    \item \textit{Defining interaction between a pair of trajectories:} We define two trajectories, $\mathcal{Y}^{GT}_i$ and $\mathcal{Y}^{GT}_j$, as interacting if the closest spatial distance between the two agents at any time point in the ground-truth (GT) future is less than a threshold distance, $d_{th} = 5\text{m}$. Mathematically, we are given trajectories $\mathcal{Y}^{GT}_i = \{p_i^1, p_i^2, ..., p_i^{T_c}\}$ and $\mathcal{Y}^{GT}_j = \{p_j^1, p_j^2, ..., p_j^{T_c}\}$. 
    $||p_i^{t_1} - p_j^{t_2}||$ denotes the Euclidean distance between point $p_i^{t_1}$ from trajectory $\mathcal{Y}^{GT}_i$ and point $p_j^{t2}$ from trajectory $\mathcal{Y}^{GT}_j$. For time points $t_1 \in \{1, ..., T_c\}$ and $t_2 \in \{1, ..., T_c\}$, $d_{ij}$ is given by:
    \begin{equation}\label{chap-6-eq:6} 
    \small
    d_{ij} = min(||p_i^{t_1} - p_j^{t_2}||) 
    \normalsize
    \end{equation}
    The two agents $i$ and $j$ are interacting if $d_{ij} < d_{th}$. This is Step-1.

    \item \textit{Classifying target agent intent:} Given a target agent, proximity does not guarantee interaction. For example: if the target agent is going straight, an oncoming agent in a different lane is unlikely to contribute to its future motion. However, if the target agent is turning left, then the oncoming agent is likely to contribute to its future motion. We thus propose a heuristic conditioned on the target agent intent to filter out agents identified by Step-1. Given the trajectory information of a target agent $i$, we classify its intent into six categories: \{Straight, Lane-Change, Right-turn, Left-turn, Right-turn-Waiting, Left-turn-waiting, Other\}. This is Step-2.

    \item \textit{Filtering agents based on target agent's intent:} In this step, we use the target agent's intent from Step-2 to filter out spatially close agents that are not interacting. We propose the following heuristic: if the target agent's intent belongs to either `Left-turn' or `Left-turn-waiting', we retain oncoming agents; otherwise, we exclude oncoming agents. In summary, this step scrutinizes the target agent's intent to ascertain whether to retain or exclude oncoming agents, effectively filtering out non-interacting agents in close proximity. This is Step-3.
\end{itemize}

\end{document}